\documentclass[journal]{IEEEtran}

\usepackage{amsmath,epsfig}
\usepackage{amsmath,amsfonts, amssymb, graphicx, latexsym, mathtools}
\usepackage{algorithm2e}
\usepackage{url,graphics,subfigure,cite,calc,psfrag,theorem}
\usepackage{times}

\ifCLASSINFOpdf
\else
\fi
\begin{document}
%
\title{Visual Multiple-Object Tracking for Unknown Clutter Rate }
%
%
%

\author{Du Yong Kim
\thanks{Du Yong Kim is with the Department
	of Electrical and Computer Engineering, Curtin University, Bentley, 6102 Australia e-mail: duyong.kim@curtin.edu.au.}
}

%
%

\markboth{This paper is a preprint of a paper submitted to IET Computer Vision. If accepted, the copy of record will be available at the IET Digital Library}%
{Shell \MakeLowercase{\textit{et al.}}: Bare Demo of IEEEtran.cls for IEEE Journals}
%



\maketitle

\begin{abstract}
In multi-object tracking applications, model parameter tuning is a prerequisite for reliable performance. In particular, it is difficult to know statistics of false measurements due to various sensing conditions and changes in the field of views. In this paper we are interested in designing a multi-object tracking algorithm that handles unknown false measurement rate. Recently proposed robust multi-Bernoulli filter is employed for clutter estimation while generalized labeled multi-Bernoulli filter is considered for target tracking. Performance evaluation with real
videos demonstrates the effectiveness of the tracking algorithm for real-world scenarios.
\end{abstract}


%
\IEEEpeerreviewmaketitle
\section{Introduction}
\label{sec:intro} Multi-object tracking is one of fundamental problems in many applications. There are abundant research works, however, it is still far from practical use. The overwhelming majority of multi-target tracking  algorithms are built on the assumption that multi-object system model parameters are known a priori, which is generally not the case in practice \cite{Mahler_book}, \cite{BarShalom}. While tracking performance is generally tolerant to mismatches in the dynamic and measurement noise, the same cannot be said about missed detections and false detections. In particular, mismatches in the specification of missed detection and false detection model parameters such as detection profile and clutter intensity can lead to a significant bias or even erroneous estimates \cite{R_CPHD_Mahler}. \newline
\indent Unfortunately, except for a few application areas, exact knowledge of model parameters is not available. This is especially true in visual tracking, in which the missed detection and false detection processes vary with the detection methods. The detection profile and clutter intensity are obtained by trial and error. A major problem is the time-varying nature of the missed detection and false detection processes. Consequently, there is no guarantee that the model parameters chosen from training data will be sufficient for the multi-object filter at subsequent frames.\newline
\indent In radar target tracking applications, stochastic multi-object tracking algorithms based on Kalman filtering or Sequential Monte Carlo (SMC) method have been widely used \cite{BarShalom}, \cite{Reid77}. This approach also has been used in visual multi-object tracking research \cite{Breitenstein11}, \cite{Tinne11}, \cite{Hoseinnezhad2013}. On the other hand, deterministic approach such as network flow \cite{DAT08}, continuous energy optimisation \cite{Milan_PAMI}, has become a popular method for multi-object tracking problem in visual tracking application. This approach is known to be free from tuning parameters, however, it is useful only when reliable object detection is available. \newline
\indent Unknown observation model parameters (i.e., clutter rate, detection profile) in online multi-object filtering was recently formulated in a joint estimation framework using random finite set (RFS) approach \cite{Mahler2003}, \cite{Mahler2007a}. Recently, Mahler \cite{R_CPHD_Mahler} showed that clever use of
the CPHD filter can accommodate unknown clutter rate and detection profile.
In \cite{Beard2013} it was demonstrated that by bootstrapping clutter
estimator from \cite{R_CPHD_Mahler} to the Gaussian mixture CPHD filter \cite{Vo2007}, performed very close to the case with known clutter parameter
can be achieved.  \cite{R_MeMBer_Vo} extended it to multi-Bernoulli filter
with SMC implementation. The multi-Bernoulli filter was used for visual multi-object tracking in \cite{DKim_ICCAIS14}. While the solution for filtering with unknown clutter rate exists, these filters do not provide tracks that identify different objects. In particular, the conference version of this work \cite{DKim_ICCAIS14} is seriously extended as a new algorithm that is able to provides track identities with completely new structure and evaluated using challenging pedestrian tracking and cell migration experiments To the best of our knowledge this paper is the first attempt for handling unknown false measurement information in online tracking. The main contribution of this paper is to design a multi-object tracker that also produces trajectories and estimates unknown clutter rate on the fly.
\section{Problem Formulation}
Let ${\mathbb{X}}=\mathbb{R}^{n_{x}}$ denote the space of the
target kinematic state, and $\{0,1\}$ denote the discrete space of labels
for clutter model and actual targets. Then, the augmented state space
is given by
\begin{equation}
	\breve{{\mathbb{X}}}={\mathbb{X}}\times
	\{0,1\}  \label{state_space}
\end{equation}
where $\times $ denotes a Cartesian product. Consequently, the state variable
contains the kinematic state, and target/clutter
indicator. We follow the convention from \cite{R_MeMBer_Vo} that the label $u=0$ will be used as a subscript to denote the clutter generators and the
label $u=1$ for actual targets.\newline
\indent Suppose that there are $T_{k}$ target and clutter object, and we have
$O_{k}$ observations (i.e., detections). In the RFS framework, the
collections of targets (including clutter objects) and measurements can be
described as finite subsets of the state and observation spaces,
respectively as (\ref{RFS_form})
\begin{equation}
	{\breve{X_{k}}} {=\{\breve{x}_{k,i}\}_{i=1}^{T_{k}}\subset \breve{{\mathbb{X}}}}{\small,}~~~~
	{Z_{k}} {=\{z_{k,j}\}_{j=1}^{O_{k}}\subset \mathbb{Z}}{\small ,}
	\label{RFS_form}
\end{equation}
where $\breve{x}_{k,i}$ represent either the kinematic state of actual target or clutter target; $z_{k,j}$ is a measurement, and $\mathbb{Z}$ is the space of measurement,
respectively. Considering the dynamic of the state, the RFS model of the
multi-target state at time $k$ consists of surviving targets and new targets
entering in the scene. This new set is represented as the union
\begin{equation}
	\begin{array}{llll}
		\breve{X}_{k}\displaystyle{=\bigcup_{\breve{x}_{k-1}\in \breve{X}_{k-1}}{S_{k|k-1}(\breve{x}_{k-1})}~~\bigcup {\Gamma _{k}}},
	\end{array}
	\label{State_Set_Propo}
\end{equation}
where ${\small {\Gamma _{k}}}$ is a set of spontaneous birth objects (actual
target or clutter targets) and ${\small {S_{k|k-1}(\cdot )}}$ is the set of
survived object states at time $k$ with survival probability $p_{S}(x)<1$.
\newline
\indent The set of observations given the multi-target state is expressed as
\begin{equation}
	\begin{array}{llll}
		Z_{k}{=Z_{T,0,k}\bigcup Z_{T,1,k}},
	\end{array}
	\label{Obs_Set_Propo}
\end{equation}
where $Z_{T,0,k}$ and $Z_{T,1,k}$ are, respectively, sets of clutter and
target-originated observations with unknown detection probability $p_{D}(x)<1
$.\newline
\indent With the RFS multi-target dynamic and measurement model, the
multi-object filtering problem amounts to propagating multi-target posterior
density recursively forward in time via the Bayes recursion. Note that in
the classical solution to this filtering problem such as PHD \cite{Mahler2003}, CPHD \cite{Mahler2007a}, and
multi-Bernoulli filters \cite{Vo_MeMber}, \cite{Hoseinnezhad2012}, \cite{Hoseinnezhad2013}, instead of clutter target measurement set, the
Poisson clutter intensity is given and the detection profile $p_{D}(x)$ is also known a priori \cite{Mahler_book}.
\section{Multi-object tracker with unknown clutter rate}
\indent The aim of this paper is to propose a new online multi-object tracker that is able to accommodate unknown clutter rate. For this purpose, the Robust Multi-Bernoulli (RMB) filter \cite{R_MeMBer_Vo} is employed to adapt unknown clutter rate. Then, the estimated clutter rate is plugged into the Generalized Multi-Bernoulli (GLMB) tracker \cite{VV13} to boost the performance in real-world scenarios.
\subsection{Robust Multi-Bernoulli Filter}
\indent The multi- Bernoulli filter parametrizes the multi-object posterior
density by using a set of pair, i.e., Bernoulli parameter, $\{(r^{(i)},p^{(i))}\}_{i=1}^{M}$ where $r^{(i)}$ and $p^{(i)}$ represent the
existence probability and the density of the state among $M$ Bernoulli
components. In the following the predicted and updated densities are represented
by propagating a set of Bernoulli parameters. The multi-Bernoulli filter
recursion for extended state space called RMB filter \cite{R_MeMBer_Vo}
is summarized to make the paper self-contained. \newline
\indent If the posterior multi-object density of the multi-Bernoulli form at
time $k-1$ is given as
\begin{equation}
	\{(r_{k-1}^{(i)},p_{u,k-1}^{(i)})\}_{i=1}^{M_{k-1}}.
	\label{previous_Bernoulli}
\end{equation}
\indent Then, the predicted intensity is approximated by the following
multi-Bernoulli
\begin{equation}
	\{(r_{k|k-1}^{(i)},p_{u,k|k-1}^{(i)})\}_{i=1}^{M_{k|k-1}}.
	\label{MB_predict}
\end{equation}
A set of predicted Bernoulli components is a union of birth components $\{(r_{\Gamma,k}^{(i)},p_{\Gamma,u,k}^{(i)})\}_{i=1}^{M_{\Gamma,k}}$ and
surviving components $\{(r_{k|k-1}^{(i)},p_{u,k|k-1}^{(i)})\}_{i=1}^{M_{k|k-1}}$. The birth Bernoulli components are chosen a priori by
considering the entrance region of the visual scene, e.g., image border. The
surviving components are calculated by
\begin{equation}
	\begin{array}{ll}
		r_{P,k|k-1}^{(i)}=r_{k-1}^{(i)}\sum_{u=0,1}\langle
		p_{u,k-1}^{(i)},p_{S,u,k}\rangle , &  \\
		p_{u,k|k-1}^{(i)}(x)=\frac{\left\langle f_{u,k|k-1}(x|\cdot,
			)p_{u,k-1}^{(i)},p_{S,u,k}\right\rangle }{\left\langle
			p_{u,k-1}^{(i)},p_{S,u,k}\right\rangle },\label{predict_eq} &
	\end{array}
\end{equation}
where $x$ is the kinematic state, $p_{S,u,k}$ is the survival probability to time $k$ and $f_{u,k|k-1}(x|\cdot)$ is the state transition density
specified by either for actual target $f_{1,k|k-1}(x|\cdot )$ or for clutter target $f_{0,k|k-1}(x|\cdot )$.\newline
\indent If at time $k$, the predicted multi-target density is
multi-Bernoulli of the form (\ref{MB_predict}), then the updated
multi-Bernoulli density approximation is composed of the legacy components
with the subscript $L$ and the measurement updated components with the
subscript $U$ as follows (\ref{update_Bernoulli})
\begin{equation}
	\begin{array}{ll}
		\{(r_{L,k}^{(i)},p_{L,u,k}^{(i)})\}_{i=1}^{M_{k|k-1}}\cup
		\{(r_{U,k}(z),p_{U,u,k}(\cdot;z))\}_{z\in Z_{k}}.\label{update_Bernoulli}
		&
	\end{array}
\end{equation}
The legacy and measurements updated components are calculated by a series of
equations (\ref{update_eq}) as follows.
\begin{equation}
	\begin{array}{lllll}
		r_{L,k}^{(i)}=\sum_{u=0,1}r_{L,u,k}^{(i)}, &  &  &  &  \\
		r_{L,u,k}^{(i)}=\frac{r_{k|k-1}^{(i)}\left\langle
			p_{u,k|k-1}^{(i)},1-p_{D,u,k}\right\rangle }{1-r_{k|k-1}^{(i)}\sum_{u^{\prime }=0,1}\left\langle p_{u^{\prime },k|k-1}^{(i)},p_{D,u^{\prime
				},k}\right\rangle }, &  &  &  &  \\
		p_{L,u,k}^{(i)}(x)=\frac{\big(1-p_{D,u,k}\big)p_{u,k|k-1}^{(i)}(x)}{\sum_{u^{\prime
				}=0,1}\left\langle p_{u^{\prime },k|k-1}^{(i)},1-p_{D,u^{\prime
			},k}\right\rangle }, &  &  &  &  \\
	r_{U,k}(z)=\sum_{u=0,1}r_{U,u,k}(z), &  &  &  &  \\
	r_{U,u,k}(z)=\frac{\sum_{i=1}^{M_{k|k-1}}\frac{r_{k|k-1}^{(i)}\big(1-r_{k|k-1}^{(i)}\big)\left\langle
			p_{u,k|k-1}^{(i)},g_{u,k}(z,|\cdot )p_{D,u,k}\right\rangle }{\big(1-r_{k|k-1}^{(i)}\sum_{u^{\prime }=0,1}\left\langle p_{u^{\prime
				}=0,1}^{(i)},p_{D,u^{\prime },k}\right\rangle \big)^{2}}}{\sum_{i=1}^{M_{k|k-1}}\frac{r_{k|k-1}^{(i)}\sum_{u^{\prime }=0,1}\left\langle
			p_{u,k|k-1}^{(i)},g_{u^{\prime },k}(z|\cdot )p_{D,u^{\prime
				},k}\right\rangle }{1-r_{k|k-1}^{(i)}\sum_{u^{\prime }=0,1}\left\langle
			p_{u^{\prime },k|k-1}^{(i)},p_{D,u^{\prime },k}\right\rangle }}, &  &  &  &
	\\
	p_{U,u,k}(x;z)=&  &  &  &\\\frac{\sum_{i=1}^{M_{k|k-1}}\frac{r_{k|k-1}^{(i)}}{1-r_{k|k-1}^{(i)}}p_{u,k|k-1}^{(i)}(x)g_{u,k}(z|x)\cdot p_{D,u,k}}{\sum_{u^{\prime
			}=0,1}\sum_{i=1}^{M_{k|k-1}}\frac{r_{k|k-1}^{(i)}}{1-r_{k|k-1}^{(i)}}\left\langle p_{u^{\prime },k|k-1}^{(i)},g_{u^{\prime },k}(z|\cdot
		)p_{D,u^{\prime },k}\right\rangle }\label{update_eq}
\end{array}
\end{equation}
where $p_{D,u,k}$ is the state dependent detection probability, $g_{u,k}(z|x)
$ is the measurement likelihood function that will be defined in the
following section. Note that the SMC implementation of summarized equations (6)-(10) can be found in \cite{R_MeMBer_Vo}.
\subsection{Boosted Generalized labeled Multi-Bernoulli Filter}
\begin{figure}
	\begin{center}
		\includegraphics[height=.55\linewidth]{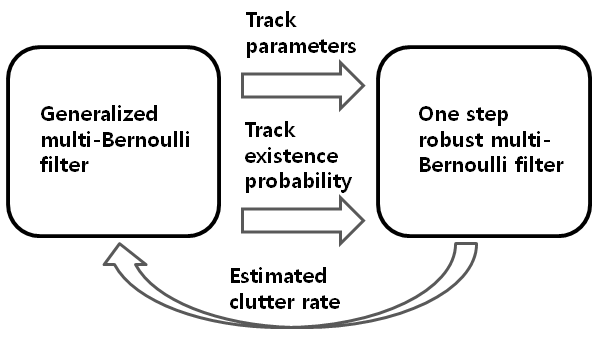} 
	\end{center}
	\caption{Schematic diagram of the proposed tracker}
	\label{Fig:Diagram}
\end{figure}

The generalized labeled multi-Bernoulli (GLMB) filter provides a solution of multi-object Bayes filter with unique labels. In this paper, the GLMB filter is used as a tracker that returns trajectories of multi-object given the estimated clutter rate from the RMB. As shown in Fig. \ref{Fig:Diagram}, GLMB and RMB filters are interconnected by sharing tracking parameters and facilitate feedback mechanism in order for robust tracking against time-varying clutter background. Note that one step RMB filter is used for the estimation of clutter rate, thus, it is not a parallel implementation of independent filter.
 
We call the proposed tracker as Boosted GLMB tracker.\newline
\indent In multi-object tracking with labels, formally, the state of an object at time $k$ is defined as $\mathbf{x}_{k}=(x_{k},\ell _{k})\in \mathbb{X\times L}_{k}$, where $\mathbb{L}_{k}$
denotes the label space for objects at time $k$ (including those born prior
to $k$). Note that $\mathbb{L}_{k}$ is given by $\mathbb{B}_{k}\cup \mathbb{L}_{k-1}$, where $\mathbb{B}_{k}$ denotes the label space for objects born at time $k$ (and is disjoint from $\mathbb{L}_{k-1}$) and we do not consider clutter generator in designing GLMB, thus, the label $u$ is omitted. Suppose that there are $N_{k}$ objects at time $k$ as (\ref{RFS_form}), but only consider actual target with label $\mathbf{x}_{k,1},...,\mathbf{x}_{k,N_{k}}$, in the context of multi-object tracking, 
\begin{equation}
\mathbf{X}_{k}=\{\mathbf{x}_{k,1},...,\mathbf{x}_{k,N_{k}}\}\in \mathcal{F}(%
\mathbb{X\times L}_{k}),
\end{equation}%
where $\mathcal{F}(\mathbb{X\times L}_{k})$ denotes the space of finite
subsets of $\mathbb{X\times L}_{k}$. We denote cardinality (number of
elements) of $\mathbf{X}$ by $|\mathbf{X}|$ and the set of labels of $\mathbf{X}$, $\{\ell :(x,\ell )\in \mathbf{X}\}$, by $\mathcal{L}_{\mathbf{X}}$. Note that since the label is unique, no two objects have the same label,
i.e. $\delta_{|\mathbf{X}|}(|\mathcal{L}_{\mathbf{X}}|)=1$. Hence $\Delta (%
\mathbf{X})\triangleq $ $\delta _{|\mathbf{X}|}(|\mathcal{L}_{\mathbf{X}}|$
is called the \emph{distinct label indicator}.\newline
\indent In the GLMB the posterior density takes the form of a
generalized labeled multi-Bernoulli 
\begin{equation}
\mathbf{\pi}_{k-1}(\mathbf{X})=\Delta (\mathbf{X})\sum_{c\in \mathbb{C}}\omega _{k-1}^{(c)}(\mathcal{L}_{\mathbf{X}})\left[ p_{k-1}^{(\xi )}\right] ^{\mathbf{X}}.
\label{generalMulti_Bernoulli} 
\end{equation}
Given the posterior multi-object density of the form (\ref{generalMulti_Bernoulli}), the predicted multi-object density to time $k$ is given by 
\begin{equation}
\mathbf{\pi}_{k|k-1}(\mathbf{X})=
\displaystyle\Delta (\mathbf{X})\sum_{c\in \mathbb{C}}\omega_{k|k-1}^{(c
	)}(\mathcal{L}_{\mathbf{X}})\left[ p_{k|k-1}^{(c)}\right] ^{\mathbf{X}}  \label{eq:PropCKstrong1}
\end{equation}
where\\
\indent \indent \indent$\omega_{k|k-1}^{(c)}(L)=w_{B,k}(L\cap \mathbb{B}_{k})\omega_{S,k}^{(c)}(L\cap \mathbb{L}_{k-1}),$\\
$p_{k|k-1}^{(c)}(x,\ell )=1_{\mathbb{L}_{k-1}}(\ell )p_{S,k}^{(c)}(x,\ell )+(1-1_{\mathbb{L}_{k-1}}(\ell ))p_{B,k}(x,\ell ),$\\
\indent \indent $p_{S,k}^{(c)}(x,\ell ) =\frac{\left\langle p_{S,k-1}(\cdot ,\ell
	)f_{k|k-1}(x|\cdot ,\ell ),p_{k-1}^{(c)}(\cdot ,\ell )\right\rangle }{\eta
	_{S}^{(c)}(\ell )},$\\
\indent $\eta_{S,k}^{(c)}(\ell ) =\int \left\langle p_{S,k-1}(\cdot ,\ell
)f_{k|k-1}(x|\cdot ,\ell ),p_{k-1}^{(c)}(\cdot ,\ell )\right\rangle dx,$\\
\indent ~~~$\omega_{S,k}^{(c)}(J)=[\eta _{S,k}^{(c
	)}]^{L_{k-1}}\sum_{I\subseteq \mathbb{L}_{k-1}}1_{I}(J)[q_{S}^{(c
	)}]^{I-J}\omega _{k-1}^{(c)},$\\
\indent \indent \indent $q_{S}^{(c)}(\ell )=\left\langle 1-p_{S,k-1}(\cdot ,\ell ),p_{k-1}^{(c)}(\cdot ,\ell )\right\rangle$, \\
where $c$ is the index for track hypothesis, $L$ is an instance of label set, $I$ is track labels from previous time step.\newline
\indent Moreover, the updated multi-object density is given by
\begin{equation}\begin{array}{ll}
\resizebox{.90\hsize}{!}{$
\mathbf{\pi}_{k|k}(\mathbf{X}|Z_{k})=
\Delta(\mathbf{X})\displaystyle\sum_{c
	\in \mathbb{C}}\sum\limits_{\theta \in
	\Theta_{k}}\omega_{Z_{k}}^{(c,\theta )}(\mathcal{L}_{\mathbf{X}})\left[ p_{k|k}^{(c,\theta )}(\cdot |Z_{k})\right] ^{\mathbf{X}} $}\label{eq:ProbBayes_strong1}
\end{array}
\end{equation}
where $\Theta_{k}$ is the space of mappings $\theta :\mathbb{L}_{k}\rightarrow \{0,1,...,|Z_{k}|\},$ such that $\theta (i)=\theta
(i^{\prime })>0~$implies$~i=i^{\prime }$, and\\
~\indent $\omega_{Z_{k}}^{(c,\theta )}(L) \propto \delta_{\theta
	^{-1}(\{0:|Z_k|\})}(L)\omega_{k|k-1}^{(c)}(L)[\eta_{Z_{k}}^{(c,\theta
	)}]^{L}, $\\
\indent \indent \indent $p_{k|k}^{(c,\theta )}(x,\ell |Z_k) =\frac{p_{k|k-1}^{(c)}(x,\ell)\psi _{Z_{k}}(x,\ell ;\theta )}{\eta _{Z_{k}}^{(c,\theta )}(\ell )},$\\
\indent \indent \indent $\eta_{Z_{k}}^{(c,\theta)}(\ell )=\left\langle p_{k|k-1}^{(c)}(\cdot ,\ell ),\psi _{Z_{k}}(\cdot ,\ell ;\theta )\right\rangle ,$\\
\indent $\psi _{Z_{k}}(x,\ell ;\theta )=\delta _{0}(\theta (\ell
))(1-p_{D,k}(x,\ell ))\\
\indent \indent \indent \indent \indent +(1-\delta _{0}(\theta (\ell )))\frac{p_{D,k}(x,\ell )g_{k}(z_{\theta (\ell )}|x,\ell )}{\kappa
	_{k}(z_{\theta (\ell )})}$\\
where $\kappa_{k}\sim\hat{\lambda_c}\mathcal{U(\mathcal{Z})}$ denotes the clutter density. $\hat{\lambda_c}$ is the estimated clutter rate from the RMB filter. Specifically, the extraction of clutter rate can be simply obtained by the EAP estimate of clutter target number as
\begin{equation}
\hat{\lambda_c}=\sum_{i=1}^{M_{k}}r^{(i)}_{0,k}p_{D,0,k}
\end{equation} 
where $r^{(i)}_{0,k}$ is the existence probability of clutter target introduced in the previous section, $p_{D,0,k}$ is the probability of detection for clutter targets, $\mathcal{U}(\mathcal{Z})$ is a uniform density on the observation region $\mathcal{Z}$.

\begin{table*}
	{\footnotesize \ }
	\par
	\begin{center}
		{\footnotesize \ 
			\begin{tabular}{|l|l|ccc|cccc|cc|}
				\hline
				\textbf{Dataset} & \textbf{Method} & \textbf{Recall} & \textbf{Precision} & 
				\textbf{FPF} & \textbf{GT} & \textbf{MT} & \textbf{PT} & \textbf{ML} & 
				\textbf{Frag} & \textbf{IDS} \\ \hline\hline
				& {Boosted GLMB} & 90.2 \% & 89.5 \% & 0.03 & 19 & 90 \% & 10 \% & 0.0 \% & 23
				& 10 \\ 
				PETS09-S2L1 & {GLMB} \cite{VV13} & 82.6 \% & 81.4 \% & 0.16 & 19 & 82.7 \% & 17.3 \% & 
				0.0 \% & 23 & 12 \\ 
				& {RMOT} \cite{RMOT} & 80.6 \% & 85.4 \% & 0.25 & 19 & 84.7 \% & 15.3 \% & 
				0.0 \% & 20 & 11 \\
				\hline\hline
				& {Boosted GLMB} & 83.4 \% & 85.6 \% & 0.10 & 10 & 80 \% & 20 \% & 0.0 \% & 12
				& 16 \\ 				
				TUD-Stadtmitte & {GLMB} \cite{VV13} & 80.0 \% & 83.0 \% & 0.16 & 10 & 78.0 \% & 22.0 \% & 
				0.0 \% & 23 & 12 \\ 
				& {RMOT} \cite{RMOT} & 82.9 \% & 86.6 \% & 0.19 & 10 & 80 \% & 20 \% & 
				0.0 \% & 10 & 16 \\
				\hline\hline				
				ETH& {Boosted GLMB} & 73.1 \% & 82.6 \% & 0.78 & 124 & 60.4 \% & 34.6 \% & 5.0
				\% & 110 & 20 \\ 
				BAHNHOF and & {GLMB} \cite{VV13} & 71.5 \% & 76.3 \% & 0.88 & 124 & 58.7 \% & 27.4 \% & 
				13.9 \% & 112 & 40 \\
				SUNNYDAY & {RMOT} \cite{RMOT} & 71.5 \% & 76.3 \% & 0.98 & 124 & 57.7 \% & 37.4 \% & 
				4.8 \% & 68 & 40 \\ \hline
			\end{tabular}
		}
	\end{center}
	\caption{Comparison with the state-of-the-art trackers}
	\label{tab:comparison_state_of_the_art}
\end{table*}

\begin{figure}
	\begin{center}
		\includegraphics[height=.80\linewidth]{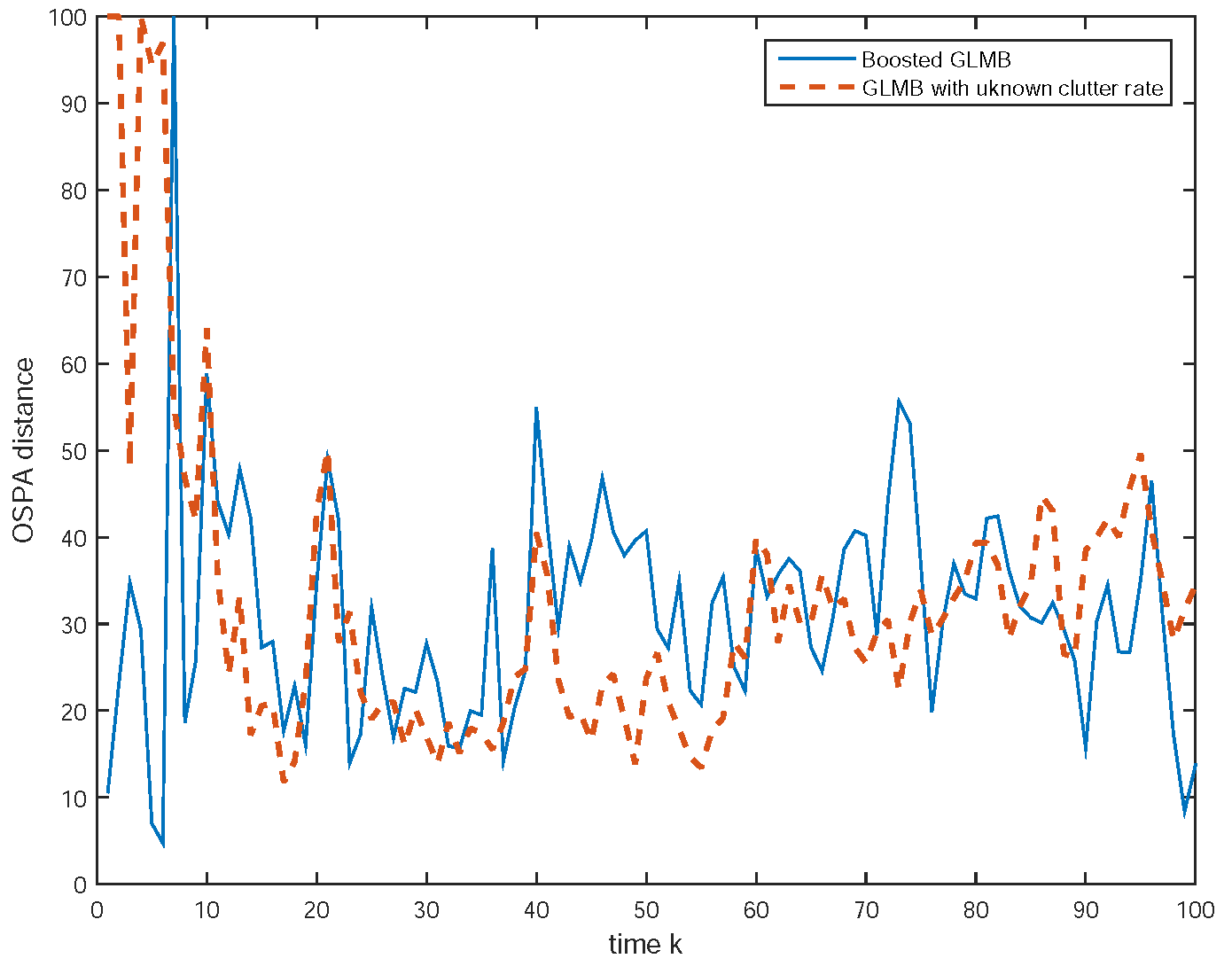} 
	\end{center}
	\caption{Comparison of OSPA distance (Boosted GLMB)}
	\label{Fig:Example1_RMB}
\end{figure}
\section{Experimental results} 
In this section, two types of experimental results are given. A nonlinear multi-object tracking example is tested in order to show the performance of the proposed tracker with respect to the standard performance metric, i.e., OSPA distance \cite{OSPA}. In addition, the proposed tracker is also evaluated for visual multi-object tracking datasets \cite{Andriluka08}, \cite{EssCVPR07}, \cite{PETS09}.
\subsection{Object motion model and basic parameters}
The target dynamic described as a coordinated turn model as (\ref{turn_model})
\begin{equation}
f_{1,k|k-1}=\mathcal{N}(x_{k};m_{x,1,k|k-1}(x_{k-1}),P_{x,1,k|k-1}),
\label{turn_model}
\end{equation}
where $m_{x,1,k|k-1}(x_{k-1})=[F(\omega_{k-1})x_{k-1},\omega_{k-1}]^T$, $P_{x,1,k|k-1}=diag([\sigma^2_wGG^T,\sigma^2_{\omega}])$,
\begin{equation}
F(\omega) =
\begin{bmatrix}
1  & \frac{\text{sin}~\omega T}{\omega} & 0 & -\frac{1-\text{cos}~\omega T}{\omega}  \\
0  &  \text{cos}~\omega T & 0 & -\text{sin}~\omega T \\
0 &  \frac{1-\text{cos}~\omega T}{\omega}  &1 & \frac{\text{sin}~\omega T}{\omega}\\
0 & \text{sin}~\omega T & 0 & \text{cos}~\omega T
\end{bmatrix},
G=
\begin{bmatrix}
\frac{T^2}{2} &0\\
T &0\\
0 &\frac{T^2}{2}\\
0 &T
\end{bmatrix},
\end{equation}
where $T$ is the sampling time, $\sigma_w$ is the standard deviation of the process noise, $\sigma_{\omega}$ is the standard deviation of the turn rate noise. These standard deviation values are determined by the maximum allowable object motion with respect to the image frame rate. For clutter targets, the transition density $f_{0,k|k-1}$, is given as a random walk to describe arbitrary motion \cite{R_MeMBer_Vo}.
\subsection{Numerical example}
The proposed algorithm is tested with a nonlinear multi-target tracking scenario in \cite{VV13}, \cite{R_MeMBer_Vo}. The actual target is observed from noisy bearing and range information $z_{k}=[\theta_{k}, r_{k}]^{T}$ and its likelihood function is given by
\begin{equation}
\begin{array}{llll}
g_{k}(z_k|x_k) = \mathcal{N}(z_k;m_{z,1,k}(x_{k}),P_{z,1,k}),
\end{array}\label{Likelihood1}
\end{equation}
where $m_{z,1,k}(x_{k})=[\text{arctan}(p_{x,k}/p_{y,k}),\sqrt{p^{2}_{x,k}+p^{2}_{y,k}}]$ and $P_{z,1,k}=diag([\sigma^{2}_{\theta},\sigma^{2}_{r}])$. For RMB implementation, we follow the same parameter setting as given in \cite{R_MeMBer_Vo}. Performance comparison between the GLMB tracker with known clutter rate and the proposed tracker (Boosted GLMB tracker) is studied. As can be seen in Fig. 1, OSPA distances for both trackers are similar. This result verifies that the Boosted GLMB shows reliable performance even when the clutter rate is unknown.

\subsection{Pedestrian tracking in vision}
\indent For the evaluation in real-world data, we are interested in tracking of multiple pedestrians. To detect pedestrians, we apply the state-of-the-art pedestrian detector proposed by Piotr et. al, called ACF detector \cite{Detect}. The detector used in the experiment integrates a set of image channels (normalized gradient magnitude, histogram of oriented gradients, and LUV color channels) to extract various types of features in order to discriminate objects from the background.\\
\indent Assuming that the object state $x_k=[p_{x,k} \dot{p}_{x,k}, p_{y,k} \dot{p}_{y,k}]^{T}$ (x-, y- positions and velocities) is observed with additive Gaussian noise, the measurement likelihood function is given by
\begin{equation}
	\begin{array}{llll}
		g_{k}(z_k|x_k) = \mathcal{N}(z_k;Hx_k,\Sigma),
	\end{array}\label{Likelihood2}
\end{equation}
where $\mathcal{N}(z;m,P)$ denotes a normal distribution with mean $m$ and covariance $P$,  $z_k$ is the response from designated detector; $H=[1~~0~~0~~0;~~1~~0~~0~~0]$ i,e., x-, y- position is observed by the detector, $\Sigma$ is the covariance matrix of the observation noise.\\
\indent Sample detection results in Fig. 2 contain false positive detections from other types of object with similar shapes as pedestrians. Based on our experiences ACF detector is robust to partial occlusions, however, there are more false positive detections than other single-model based detectors \cite{HOG}. Thus, it is relatively difficult to remove false positive detections by hard thresholding when it is used for visual scenes with time-varying imaging conditions or moving camera. In particular, in visual scene from autonomous vehicles, the average number of clutters (i.e., clutter rate) is varying with respect to the change in the field of view due to the vehicle pose change.
\begin{figure}[t]
	\begin{center}
		\includegraphics[height=.27\linewidth]{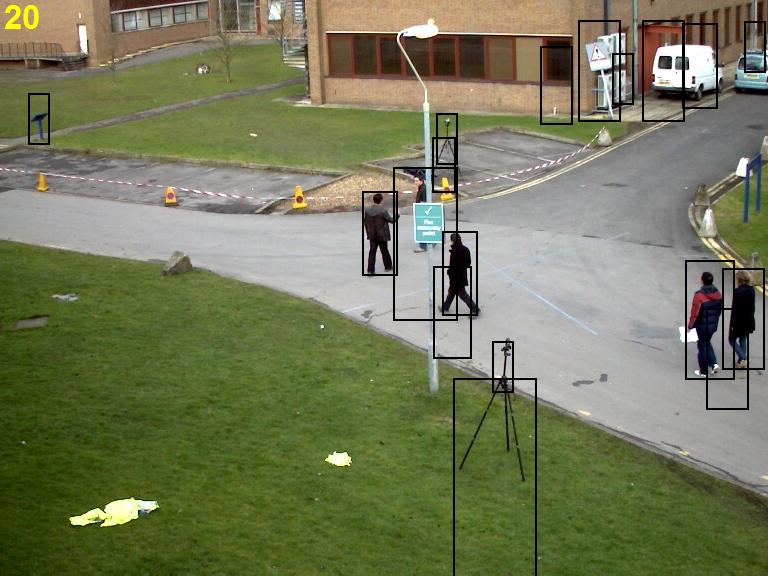}
		\includegraphics[height=.27\linewidth]{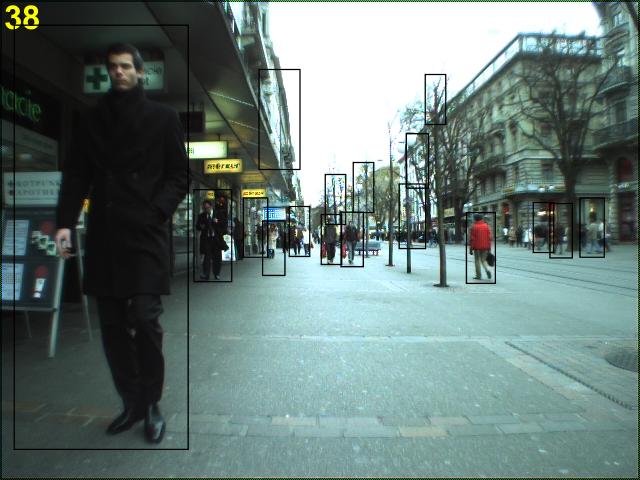}
	\end{center}
	\caption{Pedestrian detection results with clutter measurements}
	\label{Fig:clutter}
\end{figure}
The basic assumption behind the existing visual multi-object tracking is that the offline-designed detector, e.g., HOG detector \cite{Detect}, \cite{HOG} gives reasonably clean detections. Thus, direct data association algorithms such as \cite{DAT08}, and \cite{Milan_PAMI} show reasonable performance with minor number of clutter measurements. However, in practice, there are false positive detections which make data association results inaccurate and computationally intensive.\newline
\indent "S2.L1" sequence from the popular PETS'09 dataset \cite{PETS09}, "TUD-Stadtmitte" sequence from TUD dataset \cite{Andriluka08}, and "Bahnhof" and "Sunnyday" sequences from ETHZ dataset \cite{EssCVPR07} are tested in the experiment where maximum 8-15 targets are moving in the scene. The number of targets varies in time due to births and deaths, and the measurement set includes target-originated detections and clutter. In our experiments, unlike previous works, we use the ACF detector with low threshold for nonmaximum suppression so as to have less number of miss-detections but increased false alarms with time-varying rate. It is more realistic setting especially in ETHZ dataset that is recorded with frequent camera view changes. The Boosted GLMB, is compared with the original GLMB (with fixed clutter rate) \cite{VV13}, and state-of-the-art online Bayesian multi-object tracker called RMOT \cite{RMOT}. Quantitative experiment results are summarized in Table 1 using well-known performance indexes given in \cite{KuoCVPR11}. In Table 1, Boosted GLMB shows superior performance compared to the GLMB in all indexes and compatible with the recent online tracker, RMOT. The proposed Boosted GLMB outperforms other trackers with respect to FPF where tracker is able to effectively reject clutters with estimated clutter rate. On the hand, inferior performance in Fragmentation and ID switches are observed compared with RMOT because of the lack of relative motion model. \\
\indent In summary, it is verified from the experiment that the Boosted GLMB filter is effective when the clutter rate is not known a priori which is often required in real-world applications. We make experiments using unoptimized MATLAB code with Intel 2.53GHz CPU laptop. The computation time per one image frame of size 768$\times$586 is 0.2s which is reasonably suitable for real-time visual tracking application. Further improvements can be made by code optimization to speed up.
\begin{figure}
	\begin{center}
		\includegraphics[height=.70\linewidth]{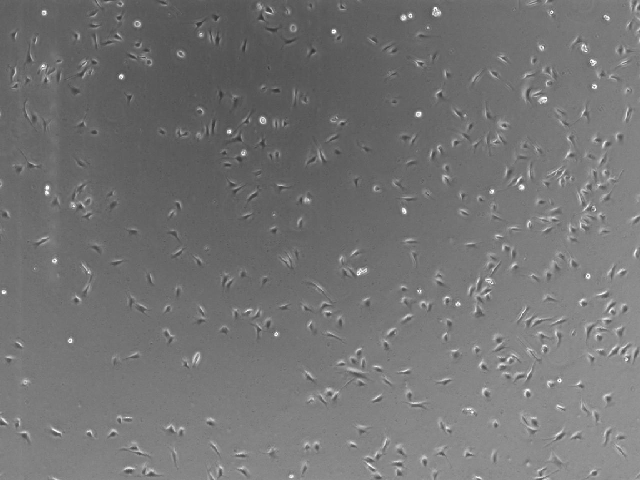}
	\end{center}
	\caption{A snapshot of microscopy image of stem cells}
	\label{Fig:cell_snapshot}
\end{figure}
\begin{figure}
	\centering
	\includegraphics[height=.41\linewidth]{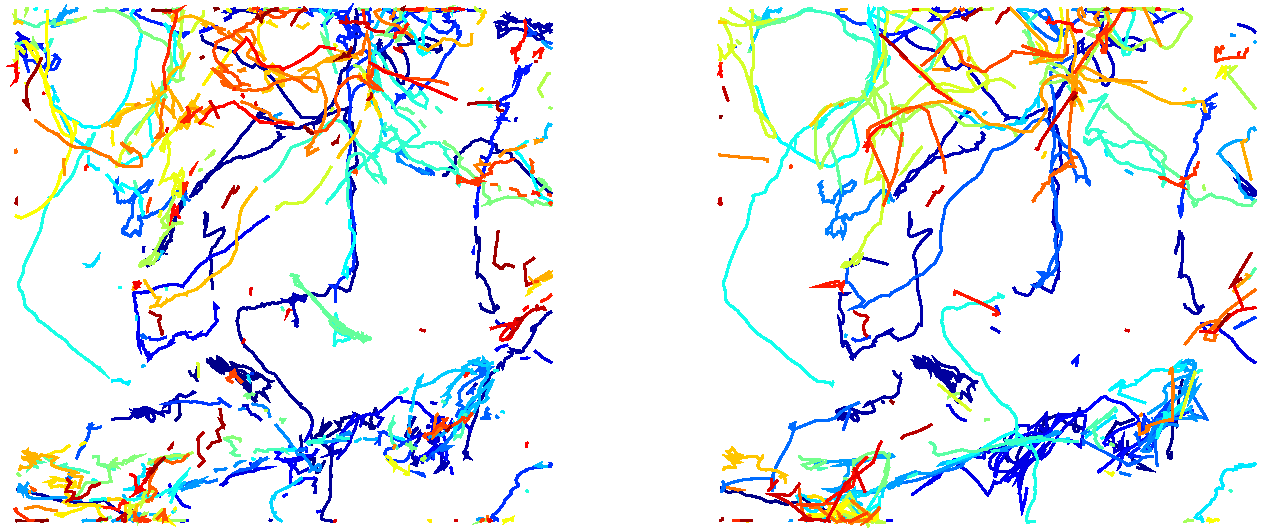}
	\caption{Reconstructed cell trajectories (left: MHT \cite{Nicolas13}, right: Boosted GLMB)}
	\label{Fig:comparision_tracks}
\end{figure}

\subsection{Cell migration analysis in microscopy image}
\begin{table}
	\caption{Comparison of averaged OSPA distance }
	\begin{center}
		{\small \ 
			\begin{tabular}{|c|c|c|}
				\hline
				\textbf{Method} & \textbf{Average OSPA}\\ 
				\hline\hline
				Boosted GLMB (Ours) & 5 \\ \hline
				MHT \cite{Nicolas13} & 8.5 \\ \hline
			\end{tabular}
		}
	\end{center}
	\label{tab:averaged OSPA}
\end{table}
The proposed algorithm is also tested with live-cell microscopy image data for cell migration analysis. The proposed GLMB tracking method is tested on a real stem cell migration sequence as illustrated in Fig. \ref{Fig:cell_snapshot}. The image sequence is recorded for 3 days, i.e., 4320 min and each image is taken in every 16 min. \newline
\indent Performance comparison is conducted with the state-of-the-art Multiple Hypothesis Tracker (MHT) \cite{Nicolas13}. The same motion and measurement models are used as in the first experiments and spot detection in \cite{Nicolas13} is applied for the fair comparison. As shown in Fig. \ref{Fig:comparision_tracks}, the Boosted GLMB provides reliable tracking results compared to the MHT. The MHT is tuned to obtain the best tracking results. The Boosted GLMB tracker produces significantly less false tracks and alleviate fragmented tracks because the tracker efficiently manages time-varying clutter information and keep confident tracks. Quantitatively, as can be seen in Table \ref{tab:averaged OSPA}, time averaged OSPA distances \cite{OSPA} for both trackers verify that the Boosted GLMB shows reliable performance even when the clutter rate is unknown.

\section{Conclusion}
\label{sec:Conclusion}
In this paper, we propose a new multi-object tracking algorithm for unknown clutter rate based on two interconnected random finite set filters. Unknown clutter rate is estimated using one step robust Bernoulli filter \cite{R_MeMBer_Vo}. Then, trajectories of objects are estimated using \cite{VV13} with estimated clutter rate online. Two filters are sharing tracking parameters so that there is no need for tuning of clutter parameters. Comparison results via a synthesized nonlinear multi-object tracking and visual tracking datasets (visual surveillance and biomedical) with state-of-the-art online trackers illustrate that the proposed multi-object tracker shows reliable performance. Interesting future research direction would be the extension of the tracking algorithm to adaptive survival probability and handling of missed-detections for further improvement.


%

\ifCLASSOPTIONcaptionsoff
  \newpage
\fi

\end{document}